\documentclass{article} % For LaTeX2e
\usepackage{rldmsubmit,palatino}
\usepackage{graphicx}
\usepackage{natbib}
\usepackage{wrapfig}
\rldmfinaltrue

\title{Task Scheduling \& Forgetting in Multi-Task Reinforcement Learning}
\author{
Marc Speckmann \\
\texttt{marc-speckmann@web.de} \\
\And
Theresa Eimer \\
Leibniz University Hannover \\
\texttt{t.eimer@ai.uni-hannover.de} \\
}

\begin{document}
\maketitle
\begin{abstract}
    Reinforcement learning (RL) agents can forget tasks they have previously been trained on. There is a rich body of work on such forgetting effects in humans. Therefore we look for commonalities in the forgetting behavior of humans and RL agents across tasks and test the viability of forgetting prevention measures from learning theory in RL. We find that in many cases, RL agents exhibit forgetting curves similar to those of humans. Methods like Leitner or SuperMemo have been shown to be effective at counteracting human forgetting, but we demonstrate they do not transfer as well to RL. We identify a likely cause: asymmetrical learning and retention patterns between tasks that cannot be captured by retention-based or performance-based curriculum strategies.
\end{abstract}

\section{Introduction}
In Reinforcement Learning (RL, \cite{sutton-book18a}), learning multiple tasks means deciding on a schedule when and how long which task will be introduced in the training process.
It has been shown that doing this in a curriculum can significantly improve the resulting policy~\citep{narvekar_curriculum_2020}.
The same is true for humans where structured repetitions in task scheduling has been shown to increase retention rates and facilitate learning~\citep{ebbinghaus_uber_1885}.
In contrast to RL approaches, however, learning theory focuses on scheduling tasks in such a way that they are repeated before they can be forgotten again.
RL curricula on the other hand focus predominantly on performance metrics~\citep{matiisen_teacher-student_2017,jiang_prioritized_2021} which will only show forgetting once it is actually occurring. 
We therefore explore the application of spaced repetition~\citep{kang_spaced_2016} principles from learning theory to task scheduling in RL by examining learning different tasks in the MiniGrid benchmark suite~\citep{chevalier-boisvert_minigrid_2023}.

First, we investigate forgetting behavior in RL generally: in learning theory for humans, we can observe so-called \textit{forgetting curves}. 
Crucially, a task is forgotten when another one is learned but with each repetition forgetting happens more slowly. 
We verify that RL shows a similar behavior in many cases.
We then apply the popular Leitner and SuperMemo systems based on forgetting curves in humans to curriculum generation in RL and compare them to Prioritized Level Replay (PLR, \cite{jiang_prioritized_2021}) which schedules tasks based on the agent's value prediction error.
We find different, though not necessarily better performance which we attribute to differences in responsiveness to scheduling. 
In fact, we do not only find such differences but see that tasks have different asymmetrical retention patterns.
We believe these results imply that effective scheduling methods would benefit from capturing the relationship between different tasks.

\section{Forgetting Curves for RL Agents}

\begin{figure}[h]
\centering
\includegraphics[width=\textwidth]{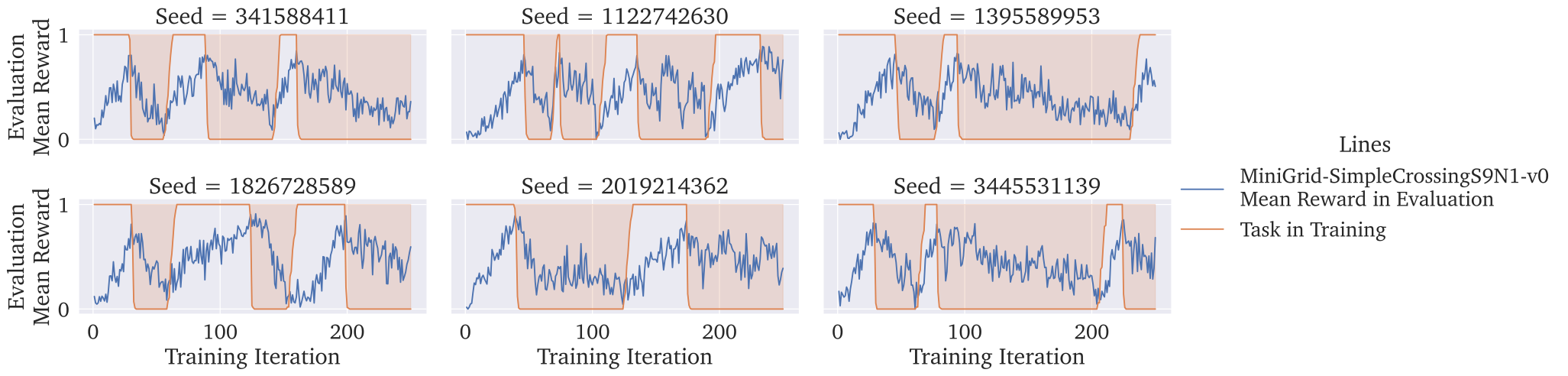}
\caption{Mean evaluation reward of the SimpleCrossing task showing decreasing forgetting}
\label{fig:decreasing_all}
\end{figure}

\begin{figure}[h]
\centering
\includegraphics[width=\textwidth]{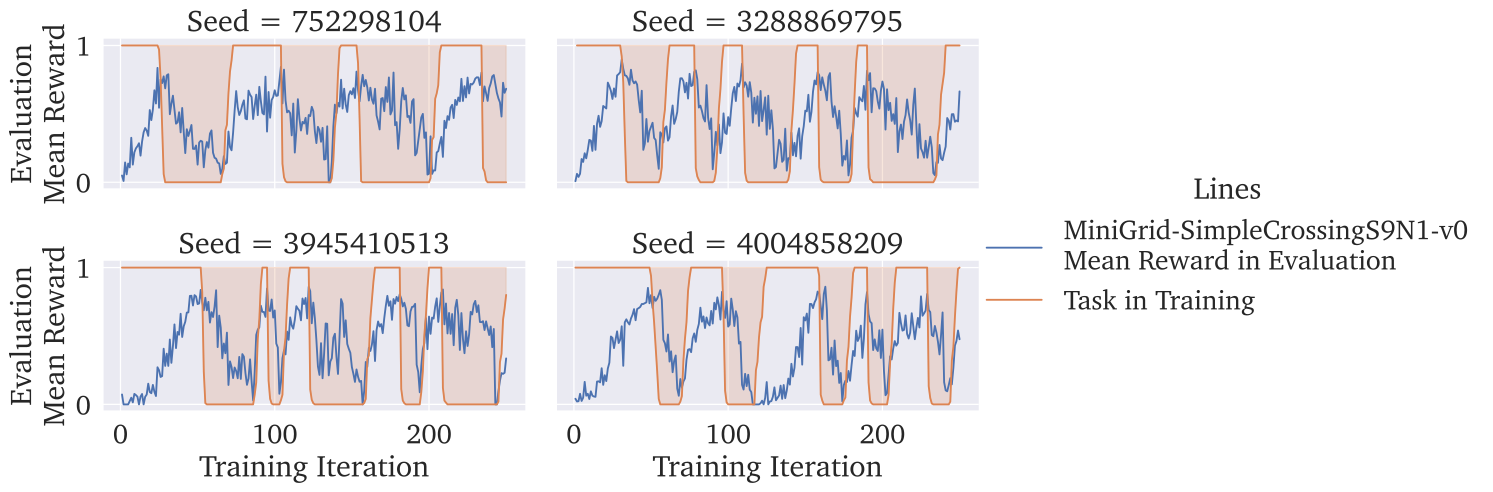}
\caption{Mean evaluation reward of the SimpleCrossing task showing periodic forgetting}
\label{fig:periodic_all}
\end{figure}

As a first step, we investigate the forgetting of a single task when learning another. 
We choose two simple tasks for this purpose: MiniGrid-Empty and MiniGrid-SimpleCrossingS9N1~\citep{chevalier-boisvert_minigrid_2023}.
We train a PPO~\citep{schulman_proximal_2017} agent on SimpleCrossing until the agent can solve it in evaluation 80\% of the time, then switch to an Empty grid until the agent only achieves a solution rate of 10\% or less.
Results are shown in Figures~\ref{fig:decreasing_all} and ~\ref{fig:periodic_all}.

In all of them, we see a distinct behavior of repeated learning and forgetting phases: the SimpleCrossing task is reliably forgotten and re-learned each time.
The speed at which this happens differs, however.
We observe two different categories of forgetting: in the six training seeds in Figure~\ref{fig:decreasing_all}, the agent behaves similarly to what learning theory tells us about human learning curves:
the task is forgotten and re-learned, but the time it takes to forget SimpleCrossing increases during training. 
The other 4 seeds we tested (see Figure~\ref{fig:periodic_all}) do not exhibit this increased retention. 
Instead, the forgetting and re-learning show a relatively regular periodic behavior.
In both cases, the agent does not forget faster over time or loses the ability to re-learn tasks. 
The fact that forgetting seems to either stay the same or slow down with repetition points to possible benefits of applying spaced repetition methods in the task scheduling process.

\section{Forgetting in Curriculum Learning}

\begin{wrapfigure}{l}{0.65\linewidth}
    \centering
    \includegraphics[width=\linewidth]{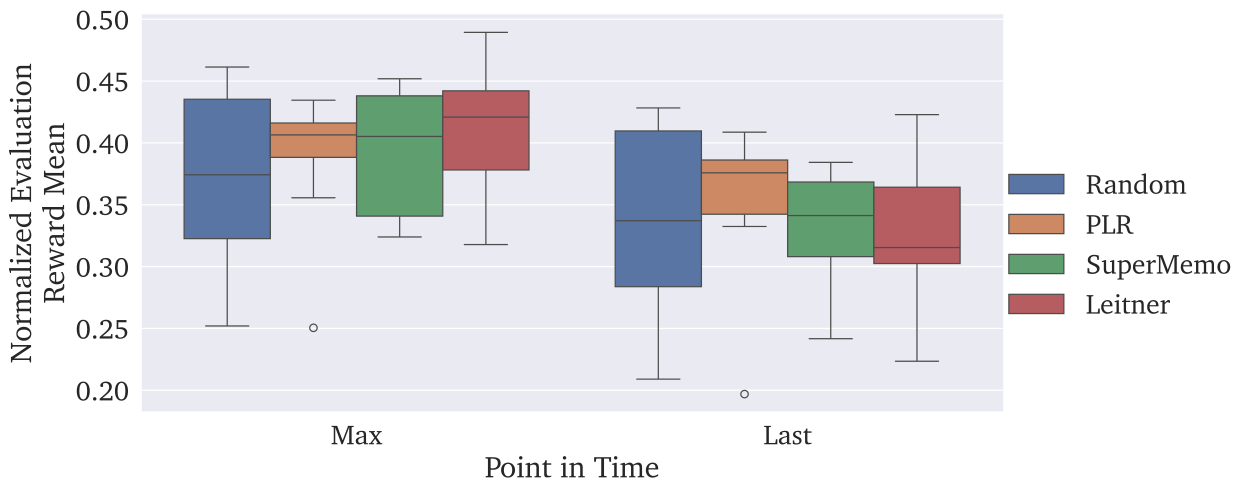}
    \caption{Normalized mean evaluation rewards of all curricula; \#runs=10}
    \label{fig:curricula_perf_comp}
\end{wrapfigure}
We integrate two popular task repetition methods from learning theory, Leitner~\citep{leitner_so_1991} and SuperMemo~\citep{wozniak_optimization_1990}, into the framework of PLR. 
We retain the existing buffer for already played levels or tasks but adapt the task selection mechanism to structured repetition.
Leitner is based on moving tasks through five stages with correctly solved tasks moving up to the next stage while incorrect ones move back to stage one.
SuperMemo on the other hand takes solution quality into account (reward in our case) and schedules tasks that are often solved badly in shorter intervals than tasks that are mostly solved well.
We use these systems to assign sampling probabilities to the tasks and compare them to the original PLR using value error and a random sampling baseline.
We use 15 tasks from MiniGrid~\citep{chevalier-boisvert_minigrid_2023} and 10 seeds per method.

\begin{figure}
    \centering
    \includegraphics[width=\textwidth]{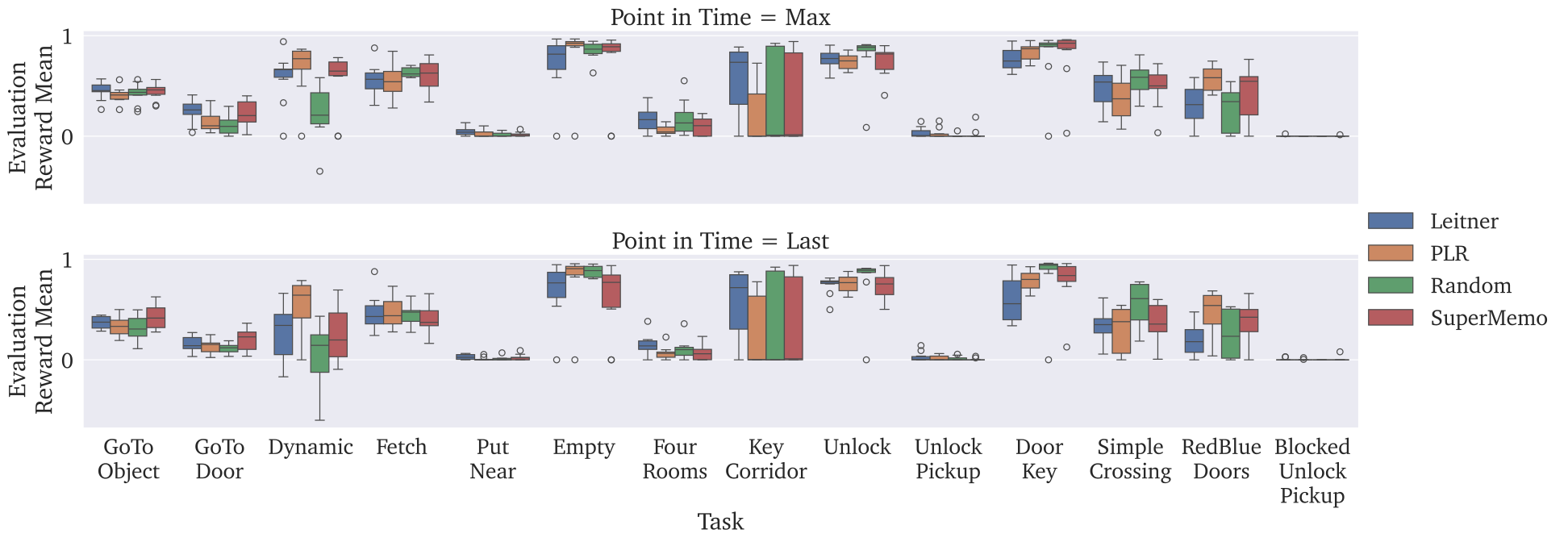}
    \caption{Mean evaluation reward of all curricula per task; \#runs=10}
    \label{fig:curricula_perf_comp_task}
\end{figure}

If forgetting is a limiting factor in PLR's performance, we would expect to see an improvement when using spaced repetition scheduling.
Figure~\ref{fig:curricula_perf_comp} shows the mean evaluation reward across the full task set for all curricula both at the maximal value we recorded and the final performance. 
As expected, the random task selection has by far the largest spread of performances but its mean is surprisingly close to the other approaches.
PLR is the most stable and achieves the highest final mean performance, though there is significant overlap between all methods.
The task repetition approaches both have a larger difference between maximum and final performance as well as a larger performance spread than PLR. 
Looking at the performance on a per-task level in Figure~\ref{fig:curricula_perf_comp_task}, some tasks like Dynamic, KeyCorridor, and RedBlueDoors exhibit far larger variation across runs in Leitner and SuperMemo than PLR.
It seems that task repetition alone is not enough for task scheduling - though neither is PLR's performance strategy since on some tasks like GoToDoor or DoorKey, SuperMemo performs much better. 
In fact, even random selection is best on some tasks like SimpleCrossing and Unlock. 
Therefore it is plausible that in this multi-task setting, both strategies, performance-based and repetition-based, are suboptimal and cannot create schedules that take all task interactions into account.

\section{Repetition Is Not Enough: Forgetting Related Tasks}
An important question for scheduling is how scheduling one task will influence another. The example above shows that SimpleCrossing is repeatedly forgotten when alternating training with Empty, it obviously does not benefit from training on Empty.
Of course, this is not true for all tasks and the idea of curricula is based on some tasks benefitting from training on others. 
Therefore we repeat this experiment with two explicitly related task pairs: Unlock and DoorKey as well as GoToDoor and GoToObject. 
We alternate training on Empty (which should again be unrelated) and one of the tasks while evaluating both.
\begin{figure}[h!]
    \centering
    \includegraphics[width=0.48\textwidth]{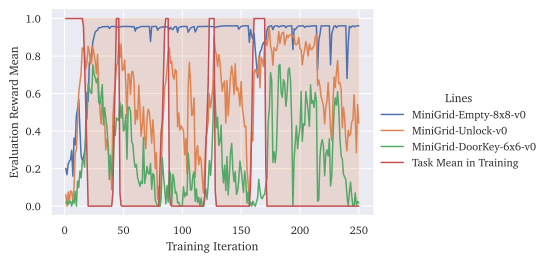}
    \includegraphics[width=0.48\textwidth]{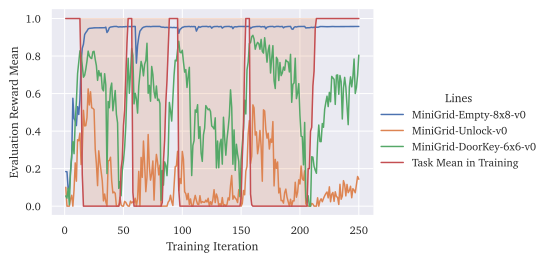}
    \label{fig:unlocks}
    \caption{Crosstraining Unlock (left) and DoorKey (right) with Empty while evaluating both.}
\end{figure}

In Figure~\ref{fig:unlocks} we see Unlock and DoorKey respectively clearly profiting from training on the other related task.
Both are learned and forgotten within the same intervals even though only one of them is trained on. 
Interestingly, however, Unlock seems to profit less from training on DoorKey than vice versa.
Thus even though the tasks are related, there is an asymmetry to the relationship: in task scheduling, we want to prioritize training on Unlock to retain this task family.

On the GoTo tasks, we see a more extreme effect. 
When training GoToObject, GoToDoor does not show a significant correlation with the learning curve while training on GoToDoor produces some improvement in GoToObject. 
Interestingly, however, while GoToDoor can positively influence GoToObject, the solution rate for both stays quite low.
Meanwhile, GoToObject is retained well and not easily forgotten when trained explicitly.
We could say a scheduling method should therefore learn GoToObject early since we can retain it easily while focusing on GoToDoor separately.

\begin{figure}
    \centering
    \includegraphics[width=0.48\textwidth]{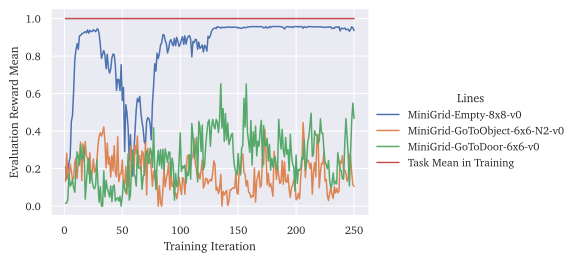}
    \includegraphics[width=0.48\textwidth]{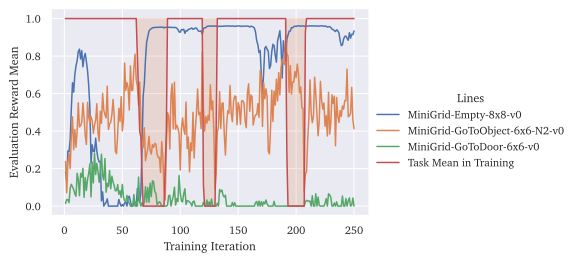}
    \label{fig:goto}
    \caption{Crosstraining GoToDoor (left) and GoToObject (right) with Empty while evaluating both.}
\end{figure}

These relationships are not captured by simple repetition, nor are tasks that the agent cannot improve upon at all (see \citet{beukman-icml24}). The same is true of purely performance-based curricula, however. 
In these results, we see that different tasks have different retention rates and asymmetric relations to one another which call for more complex scheduling behavior.
This is likely why we see differences in the performance-based and repetition-based curricula, they work well for different tasks but none works best for all of them. 
Studying and identifying such relationships on the fly could lead to improved scheduling and thus more efficient training.

\bibliography{shortstrings,shortproc,lib,local}
\end{document}